\documentclass{article}

\usepackage[utf8]{inputenc}

\usepackage[preprint]{neurips_2024}

\usepackage{placeins}
\usepackage{listings}
\usepackage[utf8]{inputenc} 
\usepackage[T1]{fontenc}    
\usepackage{hyperref}       
\usepackage{url}            
\usepackage{booktabs}       
\usepackage{amsfonts}       
\usepackage{nicefrac}       
\usepackage{microtype}      
\usepackage{xcolor}         
\usepackage{graphicx}       
\usepackage{amsmath}        
\usepackage{algorithm}      
\usepackage{algpseudocode}  
\usepackage{multirow}

\usepackage{multirow} 
\usepackage{booktabs} 

\usepackage{siunitx}  

\title{Embedding‑to‑Prefix: Parameter‑Efficient Personalization for Pre‑Trained Large Language Models}

\author{%
  \textbf{Bernd Huber,  Ghazal Fazelnia, Andreas Damianou,}\\
  \textbf{Sebastian Peleato, Max Lefarov, Praveen Ravichandran,}\\
  \textbf{Marco De Nadai, Mounia Lalmas-Roellke, Paul N. Bennett}\\
  Spotify \\
  \texttt{\{bhb, ghazalf, andreasd, speleato, mlefarov,}\\
  \texttt{praveenr, mdenadai, mounial, pbennett\}@spotify.com}
}

\begin{document}

\maketitle

\begin{abstract}
Large language models (LLMs) excel at generating contextually relevant content. However, tailoring these outputs to individual users for effective personalization is a significant challenge. While rich user-specific information often exists as pre-existing user representations, such as embeddings learned from preferences or behaviors, current methods to leverage these for LLM personalization typically require costly fine-tuning or token-heavy prompting. We propose Embedding-to-Prefix (E2P), a parameter-efficient method that injects pre-computed context embeddings into an LLM's hidden representation space through a learned projection to a single soft token prefix. This enables effective personalization while keeping the backbone model frozen and avoiding expensive adaptation techniques. We evaluate E2P across two public datasets and in a production setting: dialogue personalization on Persona-Chat, contextual headline generation on PENS, and large-scale personalization for music and podcast consumption. Results show that E2P preserves contextual signals and achieves strong performance with minimal computational overhead, offering a scalable, efficient solution for contextualizing generative AI systems.
\end{abstract}

\section{Introduction}
Large language models (LLMs) have demonstrated remarkable capabilities in generating contextually relevant content, driving significant advances across natural language processing tasks, from text completion to complex reasoning~\cite{wu2024clinicaldecoder, zhou2024difflm}. Their integration into recommendation systems has enabled content suggestions that incorporate world knowledge while maintaining coherent contextual understanding~\cite{salemi2023lamp,ma2024xrec,li2024personalizedrec}. While LLMs are proficient in general contextual understanding, effectively tailoring their outputs to reflect individual user preferences and past interactions remains a significant challenge. This is especially pertinent in recommendation systems where rich, dense user representations are learned from vast behavioral data in systems~\cite{he2017neural, fazelnia2024generalized} and offer a powerful, general-purpose source of nuanced user behavior understanding. However, effectively and efficiently integrating these dynamic, pre-computed representations to personalize LLMs, especially without costly retraining or excessive token overhead, is a major unresolved hurdle, as LLMs typically do not inherently adapt to individual user preferences~\cite{liu2025survey}.

Achieving individual personalization in LLMs is inherently challenging. Models aligned via instruction tuning or reinforcement learning from human feedback often gravitate toward majority viewpoints, making them less effective at capturing the diverse needs of individual users~\cite{santurkar2023whose}. This shortcoming is particularly evident in generative recommendation scenarios, where individual personalization consistently proves valuable~\cite{salemi2023lamp,zhang2025personalizellm,lee2023p5}. Research also suggests that user preferences are often shaped by individual, idiosyncratic values---not just broad demographic traits---highlighting the need for fine-grained personalization~\cite{hwang2023aligning}. Moreover, recent theoretical work suggests that accommodating conflicting or niche preferences within a single, generalized model may be inherently difficult~\cite{chakraborty2024maxmin}, reinforcing the case for personalized approaches. These challenges underscore the critical need for efficient personalization techniques to improve LLM performance across diverse applications.

Traditional methods for incorporating personal user context into LLMs generally fall into three categories: (1) text-based approaches, which convert user context into natural language prompts. These often add significant token overhead and may not fully leverage the information density of pre-computed vector representations~\cite{han2019transformerpersona,lee2023p5,zhang2018personachat,wu2020convai}. And (2), methods involving fine-tuning or using discrete semantic ID representations for user context. While these can capture personalization, fine-tuning is costly, and systems relying on discrete IDs (e.g., quantized embeddings) can require retraining or re-quantization when user embeddings or the underlying semantic space evolve, and may also suffer from token length issues.~\cite{kim2025prime,castricato2024persona,tang2024psw,rajput2023recommender,tan2024democratizing, jang2023personalized, madotto2019personameta}. And (3), leveraging existing user representations. While techniques for learning high-quality user representations are well-established~\cite{he2017neural, fazelnia2024generalized}, methods for effectively and efficiently integrating this pre-computed, dynamic information into pre-trained LLMs remain underdeveloped. This oversight is particularly concerning given the proven value of these dynamic user embeddings in understanding nuanced preferences.

Although methods such as Parameter-Efficient Fine-Tuning (PEFT) have emerged as a promising alternative for adapting pre-trained language models without full-scale retraining~\cite{li2021prefix,lester2021prompttuning,hu2022lora}. Most existing PEFT approaches focus on adapting LLMs to new tasks by training a small set of new parameters, rather than providing mechanisms to dynamically inject external, pre-computed rich context vectors such as user embeddings. Finally, while prompt tuning and prefix tuning have shown strong performance for task-level adaptation~\cite{li2021prefix, lester2021prompttuning, hu2022lora}, they do not provide mechanisms for directly incorporating pre-computed user embeddings into the generation process.

To address above challenges, we propose {\bf Embedding-to-Prefix (E2P)}, a parameter-efficient method that maps user embeddings directly into the LLM’s hidden representation space via prefix adapters. E2P addresses these limitations by enabling user personalization without modifying the pre-trained LLM parameters or relying on token-intensive prompt encoding of user profiles. This makes E2P particularly well-suited for large-scale, low-latency deployment scenarios. In contrast, E2P's lightweight projection module, which only requires retraining if the input embedding distribution shifts substantially, is significantly more efficient than full LLM fine-tuning or adapting LLMs to new Semantic ID spaces. In addition, continuous mapping of E2P is likely to offer greater robustness to gradual drift of embedding. E2P delivers substantial improvements across diverse tasks: dialogue personalization on Persona-Chat~\cite{zhang2018personachat}, contextual headline generation on PENS~\cite{ao2021pens}, and large-scale music and podcast recommendation.

Our main contributions are:

\begin{description}
    \item[Embedding-to-Prefix Architecture] We propose a novel method for integrating dense user embeddings directly into the hidden representation space of LLMs through a learned mapping to a single soft token prepended to the input sequence. This enables personalized generative recommendations without modifying the pre-trained LLM parameters.
    \item[Parameter-Efficient Training] E2P requires training only a small fraction of the total  model’s parameters while keeping the LLM weights frozen. This efficiency is crucial for practical deployment in production systems that must meet tight latency and cost constraints.
    \item[Validation Across Multiple Tasks] We validate our approach across four distinct tasks--dialogue personalization (Persona-Chat), contextual headline generation (PENS), and a large-scale music and podcast recommendation task, demonstrating significant gains over non-personalized baselines and strong performance compared to more token-intensive methods.
\end{description}

Our work builds upon recent advances in both preference learning for language models~\cite{huang2024lapdog,castricato2024persona} and user representation learning~\cite{liu2024personaplug,zhang2024memoryinjection}. While previous methods have explored various strategies for conditioning language models on user or community data \cite{salemi2023lamp,castricato2024persona,tang2024psw,kumar2024compo,madotto2019personameta}, our approach introduces a simple yet effective mechanism to generate personalized text using a single soft prefix token.

\section{Related Work}

\textbf{Parameter-Efficient Fine-Tuning} 
Parameter-efficient fine-tuning (PEFT) methods have emerged as powerful tools for adapting pre-trained language models to downstream tasks without requiring extensive retraining. Li and Liang~\cite{li2021prefix} introduced prefix tuning, showing that prepending continuous, task-specific vectors to transformer layers steer model behavior without updating the base model's parameters. This idea was extended by Lester et al.~\cite{lester2021prompttuning}, who demonstrated that prompt tuning can match the performance  of full fine-tuning at scale. Hu et al.~\cite{hu2022lora} proposed LoRA, which decomposes weight updates into low-rank adaptations, offering another efficient mechanism for parameter-efficient adaptation. Further PEFT approaches, such as LLaMA-Adapter~\cite{zhang2024llama}, also achieve strong performance with minimal added parameters for instruction tuning. 

However, these approaches were primarily designed for task adaptation, not personalization. That is, they learn to perform a new skill or adapt to a different data distribution, but they typically do not offer a direct mechanism for conditioning on external, pre-computed, and potentially dynamic vectors like user embeddings. Such embeddings encode instance-level context---distinct from general task adaptation---and may evolve over time, posing challenges that standard PEFT methods are not designed to address.
While existing  PEFT methods reduce training overhead and can be used to train multiple specialized or personalized models~\cite{tan2024democratizing,jang2023personalized,zhuang2024hydra}, they do not natively support external vectors, such as user embeddings, as conditioning inputs. Such dense dynamically-updated user vectors are often already available in large-scale search and recommendation settings as a key part of dense retrieval , matching, and filtering tasks.

Our proposed E2P method fills this gap by enabling the direct injection of external embedding vectors while maintaining parameter efficiency. This capability distinguishes E2P from approaches that compress textual user history or prompts into soft tokens (e.g.,~\cite{doddapaneni2024user, zhong2021useradapter, hebert2024persoma}), as E2P directly leverages \textit{pre-computed, dense} user embeddings rather than  relying on raw textual compression at inference time for personalization purposes.

\textbf{Persona and User Conditioning}
Recent work has explored various approaches to personalization in language models. Zhang et al. \cite{zhang2018personachat} introduced PersonaChat, which conditions dialogue generation on textual persona descriptions. While effective, this method requires persona attributes to be explicitly stated in natural language, consuming valuable tokens in the context window. These textual persona representations can be further optimized in retrieval-augmented setups~\cite{sun2024persona}. In a related direction, Madotto et al.~\cite{madotto2019personameta} showed that meta-learning enables rapid adaptation to new textual personas, though it still requires persona-specific fine-tuning.

More recent efforts have aimed to capture finer-grained user preferences. Chidambaram et al.~\cite{chidambaram2024direct} proposed EM-DPO, which clusters users into discrete latent preference types within an alignment framework. Kumar et al.~\cite{kumar2024compo} demonstrated significant improvements by tagging data with community identifiers via their COMPO approach. Castricato et al.~\cite{castricato2024persona} introduced the PERSONA dataset, designed for role-playing with synthetic user traits. However, these methods generally rely on textual descriptors or discrete identifiers, limiting their capacity to deliver fully individualized personalization.

In contrast, our approach repurposes existing continuous representations by  operating directly on dense user embeddings, enabling finer-grained personalization at the individual level, without relying on community-level conditioning or natural language descriptors.

\textbf{Vector-to-Text Bridges}
Several approaches have aimed to bridge the gap between vector representations and language models, often by compressing textual inputs into soft prefixes or minimal special tokens to improve efficiency~\cite{zhong2021useradapter, hebert2024persoma, li2024500xcompressor, ge2023context}. Rajput et al.~\cite{rajput2023recommender} explored semantic ID representations of entities, though their method requires computationally expensive retraining whenever user representations evolve. Balog et al.~\cite{balog2019transparent} introduced transparent semantic embeddings to connect structured data with natural language. Similarly, Liu et al. and Li et al.~\cite{liu2023visual, li2023blip} linked vision encoders to LLMs via simple linear projection layer, mapping image features into the word embedding space to support visual reasoning and instruction following. Broader efforts have also adapted LLMs to new modalities by adding specialized modules to frozen text-based backbones~\cite{shi2024llamafusion}. Related work such as ELM~\cite{tennenholtz2023demystifying} explores injecting domain-specific embeddings---including user embeddings from recommendation systems---into LLMs using trained adapters, primarily to make these embeddings interpretable. For example, USER-LLM~\cite{ning2024user} trains a dedicated user encoder from interaction sequences and integrates its (potentially multi-token) output embeddings primarily via cross-attention.

While these methods share the principle of embedding injection, E2P distinguishes itself by focusing on efficient personalized generation using a single soft prefix derived from pre-existing user embeddings, rather than on general-purpose embedding interpretation or training dedicated user encoders..
E2P enables the injection of dynamic, pre-existing embeddings through a single soft prefix with minimal impact on decoding latency. Unlike multi-modal approaches that typically map instance-specific features (e.g., image representations) into an LLM at inference time, E2P leverages continuously updated, user-specific embeddings already available in large-scale systems. By projecting these into the LLM through a lightweight prefix, E2P achieves both fine-grained personalization and computational efficiency, making it especially well-suited for production-scale deployments.

\textbf{User Representation Learning}
Dense user embeddings form the foundation of many real-world, large-scale personalized systems, particularly in content recommendation and e-commerce~\cite{he2017neural}. These systems dedicate substantial computational resources into learning  vector representations that capture complex user preferences, affinities, and behavioral signals (e.g., clicks, views, purchases, listening history). Importantly,  these embeddings are dynamic: they are continuously updated in production pipelines to reflect users' evolving tastes and recent activities~\cite{fazelnia2024generalized}, making them a timely and powerful source of personalization signals. Fazelnia et al.~\cite{fazelnia2024generalized} showed that such embeddings can effectively compress diverse aspects of user behavior into generalizable representations, achieving strong performance across a range of downstream tasks. Their proven impact on personalization metrics (e.g.~engagement and conversion) in non-generative settings~\cite{salemi2023lamp,zhang2025personalizellm,lee2023p5} underscores  their untapped potential as conditioning signals for generative models.

\section{Method}

\begin{figure}[t]
\centering
\includegraphics[width=\columnwidth]{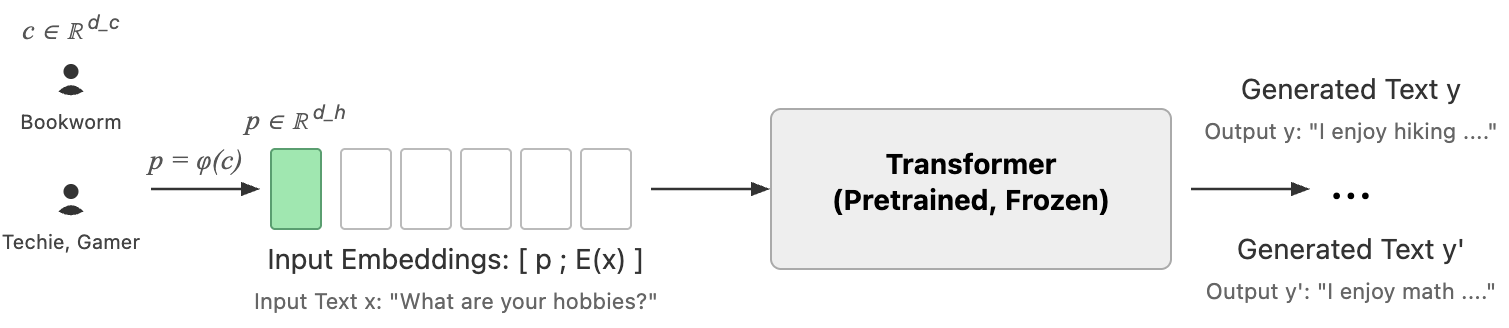}
\caption{Overview of Embedding-to-Prefix (E2P). A projection module maps a user embedding $c$ to a soft prefix token $p$ that is prepended to the LLM input embedding sequence. The soft prefix influences the entire generation process without modifying the frozen LLM weights.}
\label{fig:architecture}
\end{figure}

We introduce Embedding-to-Prefix (E2P), a parameter-efficient approach that maps user embeddings to a single soft prefix token for personalizing frozen pre-trained language models. Unlike methods that require extensive fine-tuning \cite{li2021prefix} or lengthy textual descriptions~\cite{wu2020convai}, E2P directly projects user representations into the model's hidden space. This enables efficient personalization with minimal computational overhead, while keeping the backbone model parameters completely frozen, which not only ensures efficiency but also helps mitigate complex feedback loops where behavioral signals might inadvertently alter the LLM's foundational knowledge.

\subsection{Simplified Prefix Token}
\label{subsec:prefix_tuning}

Our method builds on a simplified variant of prefix tuning~\cite{li2021prefix}, in which a trainable ``soft'' token is prepended to the input layer embeddings. This modifies the initial hidden state $\mathbf{H}_0$ as: 
\begin{equation}
    \mathbf{H}_0 = [\mathbf{p}; \mathbf{E}(\mathbf{x})]
\end{equation}
\noindent where $\mathbf{p} \in \mathbb{R}^{d_h}$ is a single learned soft token and $\mathbf{E}(\mathbf{x})$ represents the embedding of the input sequence. This design significantly reduces the number of trainable parameters and computational overhead, while still maintaining competitive performance. This aligns with prior findings that even single-token interventions can be effective~\cite{lester2021prompttuning}. 

\subsection{Projection Module}

The projection module $\phi$ maps user embeddings to the LLM's hidden space. Given a user embedding $c \in \mathbb{R}^d$, we define $\phi$ as:
\begin{equation}
\phi(c) = \text{LayerNorm}(\text{ReLU}(W_1 c)) W_2 + b
\end{equation}
where $W_1 \in \mathbb{R}^{d_c \times d}$, $W_2 \in \mathbb{R}^{d_c \times d_h}$, and $b \in \mathbb{R}^{d_h}$.

We employ a simple two-layer MLP with a ReLU activation and LayerNorm for this projection. This lightweight design was found to be effective and efficient, striking a balance between expressive power and minimizing trainable parameters, consistent with the parameter-efficient goals of E2P.

In contrast to approaches that learn static, task-specific prefixes~\cite{li2021prefix}, our projection dynamically generates personalized prefixes from continuous user embeddings. This allows the model to generalize to new users and capture similarities between users with close embeddings, leveraging the structure preserved in the prefix space (see Appendix, Figure~\ref{fig:tsne} for a visualization of how structural properties are maintained).

\subsection{Soft Token Insertion}

Given the projected vector $p = \phi(c) \in \mathbb{R}^{d_h}$, we insert it as a soft token at the beginning of the input sequence. For an input sequence $x = (x_1, \ldots, x_T)$ with embeddings $E(x) = (E(x_1), \ldots, E(x_T)) \in \mathbb{R}^{T \times d_h}$, the modified input becomes:
\begin{equation}
\hat{E}(x) = [p; E(x)] \in \mathbb{R}^{(T+1) \times d_h}
\end{equation}
where $[;]$ denotes concatenation along the sequence (time) dimension. This augmented embedding sequence is then passed to the language model.

We insert only a single token to maintain efficiency (minimal overhead), compatibility with frozen models, and overall simplicity. This design aligns with prior findings that single-token interventions can be surprisingly effective~\cite{lester2021prompttuning}. It also distinguishes E2P from more complex approach---such as prefix-tuning or LoRA methods---that modify internal representations across multiple layers~\cite{li2021prefix, hu2022lora}, and from semantic ID methods that require encoding user information via multiple hard tokens.

The E2P architecture described above---comprising the projection module $\phi$ and the soft token insertion---provides the mechanism to condition the LLM on a user context vector $c \in \mathbb{R}^{d_c}$. Formally, let $x=(x_1,\dots,x_T)$ be an input token sequence and $y=(y_1,\dots,y_S)$ the desired output. A standard autoregressive language model predicts $p_\theta(y|x)=\prod_{t=1}^{S} p_\theta(y_t|y_{<t},x)$. With E2P, the objective is to personalize this generation by modeling the conditional probability $p_\theta(y|x,c)$, where $\theta$ represents the (frozen) parameters of the pre-trained LLM. The soft prefix $p = \phi(c)$, generated from the user context, is how $c$ influences the generation. The parameters of the projection module $\phi$ are trained to optimize this conditional probability, as detailed in the following training objectives, while the LLM parameters $\theta$ remain unchanged. This ensures the parameter efficiency central to E2P.

\subsection{Training Objectives}

The training objective depends on the task. We describe the objectives used for both text generation and recommendation settings.

\subsubsection{Text Generation Tasks}

For dialogue and text generation tasks, we adopt the standard language modeling objective. Given a user embedding $c$, its corresponding soft prefix $p = \phi(c)$, and a target sequence $y = (y_1, \ldots, y_T)$, we maximize the log-likelihood:
\begin{equation}
\mathcal{L}_{LM}(c) = \sum_{t=1}^{T} \log p_{\theta}(y_t | y_{<t}, [p; x])
\label{eq:llikelihood}
\end{equation}
where $p_{\theta}$ denotes the output  distribution from the frozen language model with parameters $\theta$, and $[p; x]$ is the soft prefix concatenated with the input sequence.

\subsubsection{Recommendation Tasks}

For recommendation-oriented tasks with binary engagement labels (e.g., click/no-click), we use Kahneman-Tversky Optimization~\cite{ethayarajh2024kto}, which accounts for  both positive and negative outcomes. Given engagement labels $y_i \in \{0,1\}$, the objective is:
\begin{equation}
\mathcal{L}_{KT}(c) = \sum_{i=1}^{N} y_i \log p_{\theta}(y_i | [p; x_i]) + \alpha (1-y_i) \log (1 - p_{\theta}(y_i | [p; x_i]))
\end{equation}
Here, $\alpha$ is a weighting factor that balances false positives and false negatives, and $p_{\theta}(y_i | [p; x_i])$ is the model's predicted probability of engagement for the $i$-th example.

Compared to methods that encode user information as text~\cite{zhong2022refine, wu2020convai} (often requiring lengthy prompts) or those that fine-tune separate models per user, E2P provides a simple and scalable alternative. E2P enables personalized text generation by directly leveraging pre-existing user embeddings without incurring significant token or computational overhead.

\section{Experimental Setup}
\label{sec:experimental_setup}
To evaluate the effectiveness of our E2P method, we conduct experiments across four diverse datasets spanning personalized dialogue, personalized news headline generation, and large-scale personalization tasks.

\subsection{Datasets}
We evaluate our method on two public datasets and two large-scale production datasets, spanning both generative and recommendation tasks:
\begin{description}
\item[Persona-Chat:] This dataset~\cite{zhang2018personachat} consists of dialogues between pairs of crowdworkers, each assigned a "persona" described by 5-7 sentences. It contains 122,499 training examples, 7,801 validation examples, and 7,801 test examples. Each example include a dialogue history, a persona description, and a target response. Following prior work~\cite{zhang2018personachat}, we frame the task as next utterance prediction conditioned on dialogue history and persona. To generate embeddings, we process each persona descriptions using Llama-3.2-1B~\cite{grattafiori2024llama}, feeding the full set of persona sentences into the model, and use the last hidden state to obtain a single 2048-dimensional embedding per persona. This results in a static embedding for each unique persona description used in the dataset.
\item[ PENS:] The Personalized News Headline Corpus~\cite{ao2021pens} contains news headlines authored by professional editors for specific news articles, paired with profiles of potential readers. We use this dataset to evaluate our model's ability to generate personalized headlines based on user profiles, following the setup in~\cite{ao2021pens}. Since this corpus does not provide textual persona profiles, we construct them by using the top 20 clicked headlines (ranked by click count and dwell time), forming a behavioral profile that we encode into a vector representation using the same Llama-3.2-1B embedding method as in the Persona-Chat dataset. This process creates a static embedding for each user profile based on their historical data.
\item[Large-Scale Personalization Datasets:] We evaluate our approach on two production-scale datasets from a major streaming service.  
 {\bf Music Rec}: This task involves generating a personalized music playlist in response to a user query (e.g., ``upbeat songs for coding''). User context is provided by a proprietary 120-dimensional behavioral embedding that captures listening patterns, based on methodologies similar to~\cite{he2017neural, fazelnia2024generalized}. This embedding is dynamically updated to reflect ongoing user interactions. {\bf Podcast Rec}: This task focuses on recommending the next podcast episode for a user. As with Music Rec, personalization is achieved using a proprietary 120-dimensional behavioral embedding. Similarly, this embedding is dynamically updated. Retrieval is modeled as auto-regressive generation task, consistent with the formulation in Equation \eqref{eq:llikelihood}. In this setup, the target sequence $y$ corresponds to the semantic ID of the target entity, obtained via quantization of content's embedding space~\cite{singh2024better}. For this task, we also evaluate a textual baseline consisting of a user’s top music artists and basic demographic information. This enables a direct comparison with prompt-based personalization within this recommendation context.  In contrast, the Music Rec task primarily emphasizes the use of dense behavioral embeddings, and does not rely on readily available textual profiles.
\end{description}

\subsection{Models and Training}
We use LLaMA-3.2-3B \cite{grattafiori2024llama} as the base model for all experiments, selected for its strong performance-to-efficiency tradeoff. Throughout all experiments, the base model parameters remain frozen; only the parameters of the  E2P projection are trained.
Training is conducted using the AdamW optimizer~\cite{kingma2014adam} with a learning rate of 5e-6, a batch size of 32 and a total of 5 training epochs. All experiments were run on a machine equipped with 4xA100 80GB GPUs.

We compare E2P against the following baselines:
\begin{itemize}
\item \textbf{No Context} A vanilla LLaMA-3.2-3B decoder-only transformer with no personalization.
\item \textbf{Prompt Context} A textual description of the persona or user profile is prepended to the input as a natural language prompt, following the approach in~\cite{han2019transformerpersona} for both training and test data. While effective, this method introduces a significant token overhead. For proprietary datasets, the prompts were carefully curated to summarize key user preferences or demographics relevant to the task, aiming to match the information content of  publicly available persona where possible.
\item \textbf{E2P-Random} A control variant where the prefix is generated from a randomly selected user embedding from the test set.
\item \textbf{E2P + Prompt} Combines the E2P prefix (derived from the user embedding) with the aforementioned natural language prompt.
\item \textbf{Embedding Retrieval/Reranking} A non-generative baseline that uses the user and content embedding to retrieve or rerank candidate items based on embedding similarity. The embeddings for this baseline was computed  using a Graph Neural Network model trained on multiple content types similar to ~\cite{de2024personalized}.
\end{itemize}

Baseline applicability (detailed in Table~\ref{tab:main_results}) varies by task and data. `Prompt Context' uses textual user descriptions (available for Persona-Chat; constructed for PENS, Podcast Rec), but not for Music Rec (focused on behavioral embeddings). `E2P-Random' serves as a control, isolating the value of meaningful user embeddings from prefix structure effects. For recommendation tasks, an `Embedding Retrieval/Reranking' baseline represents a standard non-generative approach.

\subsection{Evaluation Metrics}
We employ task-appropriate metrics tailored to each dataset, recognizing that different tasks demand different evaluation criteria. Specifically, we use perplexity for language modeling (Persona-Chat), ROUGE scores for text overlap in headline generation (PENS), and engagement proxies or hit rates for recommendation effectiveness (Music and Podcast Rec).

\section{Results}
\label{sec:results}

We evaluate our Embedding-to-Prefix (E2P) method across diverse personalization tasks, comparing it against relevant baselines using task-appropriate metrics,  as summarized in Table~\ref{tab:main_results}.

Across dialogue, news, and recommendation tasks, E2P consistently outperformed non-personalized baselines and random controls, demonstrating  that projecting meaningful user embedding significantly improves generation relevance. These improvements underscore  the advantage of E2P's generative approach over traditional embedding retrieval or reranking---particularly evident in music recommendation.  In the podcast task, results further indicate that E2P's learned signal can complement textual prompts, yielding additive performance gains. 

Overall, E2P’s strong performance across multiple tasks, combined with its minimal computational overhead, highlights its practical value for scalable, personalized language generation.

\begin{table}[t]
\centering\small
\caption{Performance of E2P across multiple personalization tasks. Effects marked with $^{\dagger}$ ($^{*}$) are significant compared to No Context (Prompt-Context) baseline, while those with $^{\ddagger}$ ($\diamond$) are significant compared to E2P. Significance is determined by paired t-tests (see Appendix).
Note that we report the relative improvement (\%) compared to the No Context baseline for Music and Podcast Recommendation tasks.}
\label{tab:main_results}
\renewcommand{\arraystretch}{1.2}
\begin{tabular}{@{}c l c c c c@{}} 
\toprule
& \textbf{Method} & \textbf{Context length} & \multicolumn{3}{c}{\textbf{Metrics}} \\
\midrule

\multirow{5}{*}{\rotatebox[origin=c]{90}{\textit{Persona-Chat}}}
& & & \multicolumn{3}{c}{Perplexity} \\
\cmidrule(lr){4-6}
& No Context & 0 & \multicolumn{3}{c}{45.40} \\
& Prompt-Persona & 47 & \multicolumn{3}{c}{30.00} \\
& E2P-Random & 1 & \multicolumn{3}{c}{26.56} \\
& E2P & 1 & \multicolumn{3}{c}{\textbf{24.67$^{\dagger*}$}} \\
\midrule

\multirow{4}{*}{\rotatebox[origin=c]{90}{\textit{PENS}}}
& & & \multicolumn{1}{c}{ROUGE-1} & \multicolumn{1}{c}{ROUGE-2} & \multicolumn{1}{c}{ROUGE-L} \\
\cmidrule(lr){4-4} \cmidrule(lr){5-5} \cmidrule(lr){6-6}
& No Context & 0 & 15.64 & 5.20 & 14.66 \\
& E2P-Random & 1 & 15.57 & 5.76 & 14.79 \\
& E2P & 1 & \textbf{17.75$^{\dagger*}$} & \textbf{5.91$^{\dagger*}$} & \textbf{16.75$^{\dagger*}$} \\
\midrule

\multirow{4}{*}{\rotatebox[origin=c]{90}{\textit{Music Rec.}}}
& & & \multicolumn{1}{c}{\begin{tabular}[c]{@{}c@{}}Engagement\end{tabular}} & \multicolumn{1}{c}{\begin{tabular}[c]{@{}c@{}}Affinity\end{tabular}} & \\
\cmidrule(lr){4-4} \cmidrule(lr){5-5}
& Embedding Reranking & N/A & +0.1\% & - & \\
& E2P-Random & 1 & +5.0\% & +6.7\% & \\
& E2P & 1 & \textbf{+12.9\%$^{\dagger*}$} & \textbf{+7.2\%$^{\dagger*}$} & \\
\midrule

\multirow{6}{*}{\rotatebox[origin=c]{90}{\textit{Podcacst Rec.}}} 
& & & \multicolumn{3}{c}{Hitrate@30} \\
\cmidrule(lr){4-6}
& Embedding Retrieval & N/A & \multicolumn{3}{c}{+1.2\%} \\
& E2P & 1 & \multicolumn{3}{c}{+2.2\%$^{\dagger*}$} \\
& Prompt-Context & 41 & \multicolumn{3}{c}{+10.3\%$^{\dagger*}$} \\
& E2P-Random & 1 & \multicolumn{3}{c}{+0.7\%} \\
& E2P + Prompt & 42 & \multicolumn{3}{c}{\textbf{+13.7\%$^{\ddagger \diamond}$}} \\
\bottomrule
\end{tabular}
\end{table}

\subsection{Analysis of Results}

On the Persona-Chat dataset, E2P achieves the best perplexity score (24.67) while incurring minimal token overhead. The comparison with E2P-Random confirms that the performance gains are driven by the semantic content in the embeddings, not simply introducing  additional parameters~\cite{li2021prefix, hosseini2021empathetic}.

On the PENS personalized news headline generation task, E2P again demonstrated its effectiveness, achieving a ROUGE-L score of 16.75, which is a 14.3\% improvement over the \texttt{No Context} baseline's 14.66. The \texttt{E2P-Random} variant, which uses a prefix derived from a randomly selected user's embedding, yielded a ROUGE-L of 14.79 on PENS, performing very similarly to \texttt{No Context}. While on some other tasks (like Persona-Chat and Music Recommendation, as discussed elsewhere), \texttt{E2P-Random} can provide a noticeable improvement over a \texttt{No Context} approach --- potentially due to the prefix architecture itself offering beneficial general conditioning or the projection of even a non-specific user embedding acting as a useful, albeit generic, signal --- the PENS results highlight a scenario where such non-specific effects are minimal.

For production-scale music recommendation, E2P delivers  a 12.9\% improvement in predicted user engagement and a 7.2\% increase in user-track affinity over the non-personalized baseline. In contrast, the E2P-Random variant yields only  a 5.0\% engagement gain, confirming that E2P's benefits stem from genuine personalization rather than incidental prefix effects~\cite{yan2020emotionalpersona, song2020multistage}.

In the podcast recommendations task, E2P improves Hitrate@30 by 2.2\% over the non-personalized baseline with the addition of just a single token, and orders of magnitude fewer training parameters. While the Prompt-Context baseline achieves a higher gain (10.3\%), it requires 41 additional tokens per user, adding to sequence length and inference overhead. Importantly, combining E2P with Prompt-Context yielded the best result (+13.7\% improvement), suggesting E2P's dense embedding provides complementary personalization signals to textual prompts, capturing nuances not easily articulated in text~\cite{niknam2022personalized}.

Across all four tasks--dialogue, news, music, and podcasts--E2P consistently improves performance, demonstrating strong generalizability and practical applicability for a wide range of personalization tasks. These gains across multiple tasks are particularly impressive given the method’s minimal architectural changes---requiring only a single added token---in contrast to task-specific methods that often involve extensive customization or model fine-tuning~\cite{fan2023stylept}.

\section{Discussion and Conclusion}

In this work, we introduce Embedding-to-Prefix (E2P), a parameter-efficient method that maps dense user embeddings to a single soft token for personalizing pre-trained LLMs. Our method requires training only a fraction of the parameters of the base model while leveraging existing personalization infrastructure, making it highly practical for large-scale personalization applications. E2P demonstrates that effective personalization of LLMs can be achieved with minimal computational overhead, without modifying the backbone model, while maintaining strong performance across diverse tasks.

Our empirical evaluation shows consistent and significant improvements over non-personalized baselines: on Persona-Chat for dialogue generation, E2P achieves lower perplexity; on PENS, it improves ROUGE-L by +14.3\%; and in production-scale settings, it yields up to +12.9\% in predicted user engagement for music recommendation and +2.2\% in a podcast recommendation. These gains across multiple tasks underscore E2P’s strong generalizability, achieved with minimal architectural changes compared to task-specific methods that often require extensive model modification~\cite{fan2023stylept}.

The success of E2P validates our key hypothesis: that direct projection of dense user embeddings into the hidden space of frozen LLMs enables effective, scalable personalization while preserving computational efficiency, an essential requirement for real-world deployment. By integrating existing user embeddings directly, rather than encoding them through token-heavy text prompts or semantic ID schemes, E2P avoids the inefficiencies and information loss associated with these alternatives\cite{han2019transformerpersona,rajput2023recommender}.

While our results are promising, several limitations remain. First, E2P’s effectiveness depends heavily on the quality and relevance of the input user embeddings. While E2P is applicable to embeddings from textual descriptions (public datasets), it excels with rich, pre-existing behavioral embeddings common in production. Access to model internals for prefix injection is needed, posing a challenge for API-based models. Future work might explore distilling the E2P signal into prompts or discrete token sequences for such models, though maintaining direct injection's full fidelity is difficult. Our findings with `E2P + Prompt' suggest promising avenues for future research in combining dense embedding-based personalization with other contextual signals, such as textual descriptors or semantic IDs, to create even richer and more nuanced user representations.

Future work could pursue several additional promising directions. Extending E2P to multi-modal personalization tasks (e.g., aligning user embeddings across text, audio, and images \cite{li2024personalizedrec}) could open up new applications in cross-modal generation. Our results also suggest potential in combining multiple personalization signals—as demonstrated in our E2P + Prompt Context setting \cite{li2024personalizedrec}—which may offer richer user modeling than any single signal alone. Exploring adaptive or hierarchical prefix structures that incorporate temporal or contextual information could also enhance flexibility.

The broader implications of this work extend beyond text generation. By offering an efficient mechanism to inject personalization into frozen, large-scale foundational models, E2P contributes to a growing body of research aimed at making general-purpose AI systems more adaptable to individual users~\cite{zhong2022refine}. Our approach offers a scalable and production-ready pathway for incorporating user-specific information into LLM outputs without incurring the computational and storage cost of full model fine-tuning~\cite{li2021prefix}. 
This comes, however, with the important need to address associated societal risks, such as echo chambers and manipulative profiling.

In summary, E2P establishes that directly projecting user embeddings into the hidden states of LLMs provides an effective and computationally efficient solution to personalization—preserving the generative power of large models while enabling nuanced, user-aware outputs. This integration represents an important step toward truly personalized generative AI systems at scale.

\bibliography{references}
\bibliographystyle{plain}
\newpage
\appendix

\section{Music Recommendation: Background}

The large-scale music recommendation task involves generating playlists for user queries that often require understanding concepts beyond simple genre or artist matching, frequently termed "world knowledge" queries (e.g., "upbeat songs for coding," "music like artist X but less known"). While a base LLM can interpret the query's theme, the generated tracklist might lack resonance if it ignores the user's specific affinities within that theme (e.g., preference for electronic vs. instrumental coding music, or familiarity level with suggested artists). The core challenge is thus to steer the LLM's generation not just by the query's explicit request, but also by the user's implicit tastes captured in their behavioral embedding, aiming for playlists that feel personally curated.

Engagement Predictor Details: To evaluate the effectiveness of personalization in this offline setting, we utilized a dedicated engagement prediction model. This binary classifier was trained on a large historical dataset of user interactions with previously recommended playlists, learning to predict the likelihood of a positive engagement event (specifically, the user saving the generated playlist) given the user's embedding, the original query text, and representations of the tracks in the generated playlist. This predictor serves as a proxy metric for user satisfaction, allowing us to estimate the impact of different personalization strategies like E2P on downstream user behavior before deploying them online.

\section{Podcast Recommendation: Background}

The podcast recommendation scenario focuses on predicting the subsequent podcast a user might listen to, often leveraging the immediate context of recently played podcasts. Unlike playlist generation, this task typically emphasizes sequential prediction and ranking a single best item. The vastness of podcast content—spanning diverse topics, formats, hosts, and production styles—makes personalization particularly critical. A user's choice for the "next" podcast can be influenced by factors like narrative continuity within a series, topical relevance to recent listens, preference for specific hosts or guests, or alignment with broader, long-term interests not immediately obvious from the last few plays. E2P's application here tests its ability to inject nuanced, embedding-derived user preferences into this sequential prediction task, complementing context derived from recent listening history or explicit textual user profiles, aiming to improve the hit rate of relevant suggestions within the user's feed.

\section{Music Recommendation: Embedding Structure}
\begin{figure}[H]
    \centering
    \includegraphics[width=0.9\textwidth]{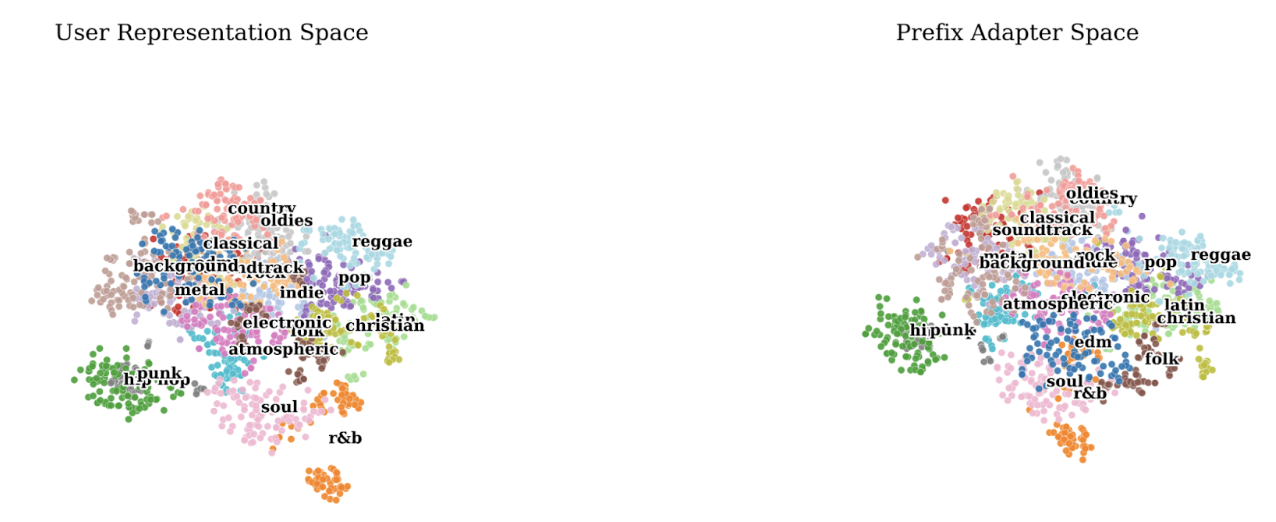}
    \caption{Visualization of user embedding properties in the LLM's hidden representation space: (a) t-SNE visualization of user embeddings colored by music genre preference clusters, (b) Corresponding visualization of the same users in prefix adapter space, demonstrating preservation of preference clusters with enhanced separation between distinct user groups. This plot demonstrates that the learned projection preserves meaningful user preference structures (e.g., genre clusters) when mapping from the user embedding space to the prefix adapter space, supporting the potential for generalization based on user similarity.}
    \label{fig:tsne}
\end{figure}

Figure~\ref{fig:tsne} demonstrates how E2P effectively preserves user preference signals in the prefix adapter space for our music recommendation task, enabling the model to generate recommendations that respect nuanced user preferences while leveraging the world knowledge capabilities of the LLM.

\section{Hyper-parameters}
\begin{table}[h]
\centering\small
\begin{tabular}{lccccc}
\toprule
Task & Base model & lr & batch & epochs & seed\\ \midrule
Persona-Chat & Llama-3.2-3B & $5{\times}10^{-6}$ & 64 & 3 & 42\\
PENS        & Llama-3.2-3B & $5{\times}10^{-6}$ & 32 & 3 & 42\\
Music Rec  & Llama-3.1-8B & $5{\times}10^{-7}$ & 256 & 5 & 42\\
Podcast Rec     & Llama-3.2-1B & $1{\times}10^{-5}$ & 16  & 5 & 42\\
\bottomrule
\end{tabular}
\end{table}

\section{Statistical tests}
\label{sec:significance}
We performed statistical significance testing to validate the improvements reported in Table~\ref{tab:main_results}. We used paired two-sided t-tests, comparing the performance metrics on the same test set instances across different methods, with a significance level of $\alpha = 0.05$. Paired tests were used because model performance was evaluated on the identical test set instances for each comparison, allowing us to control for instance-specific variations.

\section{Datasets and Splits}
\label{sec:data}

We study four corpora, two public and two proprietary.  
For each corpus we report (i) the raw source, (ii) the filtering rule, and (iii) the deterministic split.  
Code to reproduce our data splits are provided in the next section.

1.~\textbf{Persona-Chat.}  
We start from the ``cleaned'' HuggingFace edition (version dated 2024-03-01, 122{,}499 dialogues).  
We keep every triple
(<$user_0$, bot$_0$, user$_1$>) that contains exactly 3 exchanges, yielding 62,223 instances.  
The data are shuffled with seed 42 and partitioned 80/10/10 by \textit{persona id}, ensuring no persona appears in more than one split.\footnote{%
  Raw counts: train 49{,}778, dev 6{,}222, test 6{,}223.  
}
The accompanying persona string is truncated to 1{,}000 characters and is embedded once with Llama-3.2-1B in \texttt{embed} mode.

2.~\textbf{PENS.}  
We use the official v7 TSV dump distributed on Kaggle and respect its \texttt{train}/\texttt{valid}/\texttt{personalized\_test} partition.  
From every impression we create up to three positive and three negative
($\langle$)body,~headline$\rangle$) pairs (\textit{see next section for the exact loop}).  
Each user profile string is the full click history concatenated ((<)1{,}000 chars) and embedded once; the final table sizes are
train 1,317,966,
dev 2,266.
No news ID leaks across splits.

3.~\textbf{Music Playlist Generation.}  
This internal dataset consists of 300,000 user queries and generated playlists.  
Each row is labeled with an engagement annotation by the listener ((50\%/50\%)).
1,000 queries were randomly sampled chosen as test set. Every user row is joined with the latest 120-d behavioral embedding already computed in production.

4.~\textbf{Next Podcast Recommendation.}  
This internal dataset consists of 2,000,000 pairs of previously listened podcasts, metadata, and user context, and a testset of 10,000 samples. Labels are single semantic IDs; evaluation uses hit-rate@30.

\subsection{Evaluation Metrics}
Additional information on evaluation metrics by task:

\begin{description}
\item[Persona-Chat:] Following the original evaluation setup~\cite{zhang2018personachat}, we report perplexity on the test set to assess language modeling quality.
\item[PENS:] For the news headline generation task, we report ROUGE-1, ROUGE-2, and ROUGE-L scores~\cite{lin2004rouge}, which measure word overlap, bigram overlap, and longest common subsequence between generated headlines and reference headlines.
\item[Music Rec:] We evaluate this task using a binary engagement classifier trained on historical user interaction data. The classifier takes as input a user prompt and a set of track embeddings and predicts the binary signal whether or not a user engages (Improvements in this offline proxy have correlated with online gains in user satisfaction in related production experiments).
\item[Podcast Rec:] We evaluate this task using temperature sampling with $t=1.0$, computing hitrate@30. For each query, we sample 30 candidates and measure whether the true positive appears in the top 30 ranked results.
\end{description}

\paragraph{Implementation details.}
For all corpora we lower-case, de-duplicate whitespace, and preserve Unicode.  
Context embeddings all use Llama-3.2 (1B for embedding).  

\section{Pre-processing scripts}

\subsection{Persona-Chat}
\begin{lstlisting}[basicstyle=\scriptsize\ttfamily, columns=fullflexible]
import json, random, datasets, torch, numpy as np
ds = datasets.load_dataset("AlekseyKorshuk/persona-chat", "cleaned")["train"]
random.seed(42); np.random.seed(42); torch.manual_seed(42)
rows = [e for e in ds if len(e["utterances"][1]["history"]) == 3]
random.shuffle(rows)
n = len(rows); tr = int(0.8 * n); dv = int(0.9 * n)
splits = {"train": rows[:tr], "dev": rows[tr:dv], "test": rows[dv:]}
for name, data in splits.items():
    with open(f"personachat_{name}.jsonl", "w") as f:
        for r in data:
            persona = " ".join(r["personality"])[:1000]
            h = r["utterances"][1]["history"]
            prompt = f"<|system|>\n{persona}\n<|eot_id|>\n<|user|>\n{h[0]}"
            prompt += f"\n<|eot_id|>\n<|model|>\n{h[1]}"
            f.write(json.dumps({"uservector": [0.]*2048, "text": prompt})+"\n")
\end{lstlisting}

\subsection{PENS}
\begin{lstlisting}[basicstyle=\scriptsize\ttfamily, columns=fullflexible]
import json, pandas as pd, random, numpy as np, torch
P = "~/.cache/kagglehub/datasets/divyapatel4/microsoft-pens"
P += "-personalized-news-headlines/versions/7/PENS"
news = pd.read_csv(f"{P}/news.tsv", sep="\t").set_index("News ID")
train = pd.read_csv(f"{P}/train.tsv", sep="\t")
random.seed(42); np.random.seed(42); torch.manual_seed(42)
out = []
for _, r in train.iterrows():
    for lab, bit in [(r.pos, 1), (r.neg, 0)]:
        for n in lab.split()[:3]:
            b = news.loc[n]["News body"]; h = news.loc[n]["Headline"]
            p = f"<|user|>\nWrite me the title of the following news article:"
            p += f"\n{b}\n<|eot_id|>\n<|model|>\n"
            out.append({"uservector": [0.]*2048, "prompt": p,
                       "completion": h, "label": bit})
with open("pens_train.jsonl", "w") as f:
    for o in out: f.write(json.dumps(o) + "\n")
\end{lstlisting}

\end{document}